\documentclass{article} %

\usepackage{colm2024_conference}

\usepackage{booktabs}
\usepackage{graphicx}
\usepackage{enumitem}
\usepackage{wrapfig}
\usepackage{algorithm}
\usepackage{algpseudocode}
\usepackage[T1]{fontenc}

\usepackage{lmodern}
\usepackage{pifont}

\usepackage{microtype}
\usepackage{amsmath}
\usepackage{colortbl}
\usepackage[utf8]{inputenc}
\definecolor{lightgray}{rgb}{0.9,0.9,0.9}
\usepackage{caption}
\usepackage{subcaption}
\usepackage{xcolor}
\usepackage{setspace}
\usepackage{url}
\usepackage{multirow}
\usepackage{colortbl}
\usepackage{tabularx}
\usepackage{blindtext}
\usepackage{pgfplots}
\pgfplotsset{compat=1.18} 
\usepackage{tikz}
\usetikzlibrary{er,positioning,bayesnet}
\usepackage{makecell}
\usepackage{tipa}
\usepackage{siunitx}
\usepackage{nicefrac}
\usepackage{tocloft}
\usepackage{listings}
\usepackage[raster,skins]{tcolorbox} %
\usepackage{xltabular}
\usepackage{adjustbox}
\usepackage{xurl}
\usepackage{marvosym}
\usepackage{algpseudocode}

\author{\textbf{Mason Kadem}\textsuperscript{\Letter}$^{1}$ and \textbf{Rong Zheng}\textsuperscript{\Letter}$^{1}$, 
\\$^1$Computing and Software, Faculty of Engineering, McMaster University }

\title{Interpreting Transformers Through Attention Head Intervention}

\begin{document}

\maketitle

\begin{abstract}
Neural networks are growing more capable on their own, but we do not understand 
their neural mechanisms. Understanding these mechanisms' decision-making processes, 
or \textit{mechanistic interpretability},  enables (1) accountability and control in high-stakes domains, (2) the study of digital brains and the emergence of cognition, and (3) discovery of new knowledge when AI systems outperform humans. This paper traces how attention head intervention emerged as a key method for causal interpretability of transformers. The evolution from visualization to intervention represents a paradigm shift from observing correlations to causally validating mechanistic hypotheses through direct intervention. Head intervention studies revealed robust empirical findings while also highlighting limitations that complicate interpretation. Recent work demonstrates that mechanistic understanding now enables targeted control of model behaviour, successfully suppressing toxic outputs and manipulating semantic content through selective attention head intervention, validating the practical utility of interpretability research for AI safety.

\end{abstract}

\section{Introduction}

Neural networks are growing more capable on their own, but we do not understand 
their neural mechanisms~\citep{sharkey2025openproblems}. Understanding these mechanisms' decision-making processes, 
or \textit{mechanistic interpretability}, enables (1) accountability and control in high-stakes domains~\citep{rudin2019stop}, (2) the study of digital brains and the emergence of cognition, and (3) discovery of new knowledge when AI systems outperform humans.

Understanding why transformer models solve problems no human can while failing tasks three-year-olds master demands distinguishing robust mechanisms from brittle shortcuts. Concretely, mechanistic interpretability makes these causal claims. Attention head intervention achieves this by actively manipulating individual attention heads to establish which causally drive behaviour, unlike other interpretability methods that observe correlations, representing a paradigm shift toward causal understanding in transformer interpretability.

\subsection{Taxonomy}

\begin{figure}[htbp]
    \centering
    \includegraphics[width=0.9\textwidth]{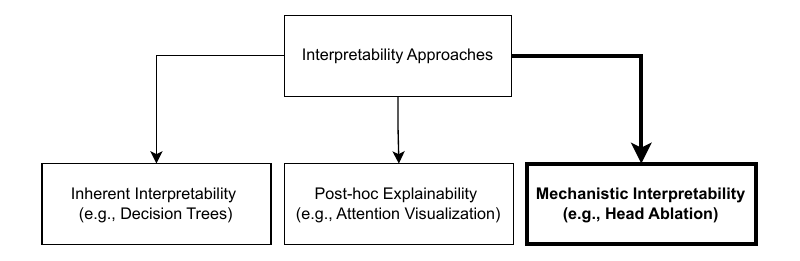}
    \caption{A high-level taxonomy of interpretability approaches. We note several important caveats. These categories are not mutually exclusive, and the same technique may serve different purposes depending on application. For example, attention visualization can be purely observational or part of mechanistic analysis. Techniques can also be distinguished by explanation type. Attention head ablation provides counterfactual explanations (what would happen without this head), while attention visualization primarily provides factual explanations (what did the model attend to), though it can be used counterfactually when comparing patterns across interventions. This paper focuses on attention head intervention (mechanistic) while examining attention visualization (observational) to trace the debate about faithful explanations.}
    \label{fig:taxonomy}
\end{figure}

The conceptual heterogeneity within interpretability research necessitates adopting a clear taxonomy. Herein, we situate mechanistic interpretability as a field distinct from both inherent interpretability and post-hoc explainability (Figure~\ref{fig:taxonomy}). 

Inherently interpretability refers to models that are interpretable by design where the structure of the model is the explanation. These models sacrifice representational capacity for transparency. Their decision processes are directly accessible to humans. For example, Decision trees make predictions through explicit rule sequences visible in tree structure. Linear models provide interpretability through weighted feature combinations where each coefficient's contribution is transparent. It prioritizes transparency over performance. The price for guaranteed interpretability is reduced capability, but not always~\citep{kadem2025sleep, kadem2023interpretable, kadem2023xgboost}

The second approach develops post-hoc explanation methods for trained models. Rather than constraining model architecture, this approach allows complex designs. Explanations are generated afterwards. These methods explain individual predictions through local approximations or attribution techniques. For example,  SHAP \citep{lundberg2020local} assigns contribution scores based on cooperative game theory. Credit for predictions is distributed across input features according to Shapley values. This approach accepts that models may be complex internally but explains specific behaviours to users. Post-hoc methods enable interpretability for powerful models while maintaining their performance advantages. However, these explanations describe behaviour without necessarily revealing underlying mechanisms.

\begin{figure}[htbp]
    \centering
    \includegraphics[width=\textwidth]{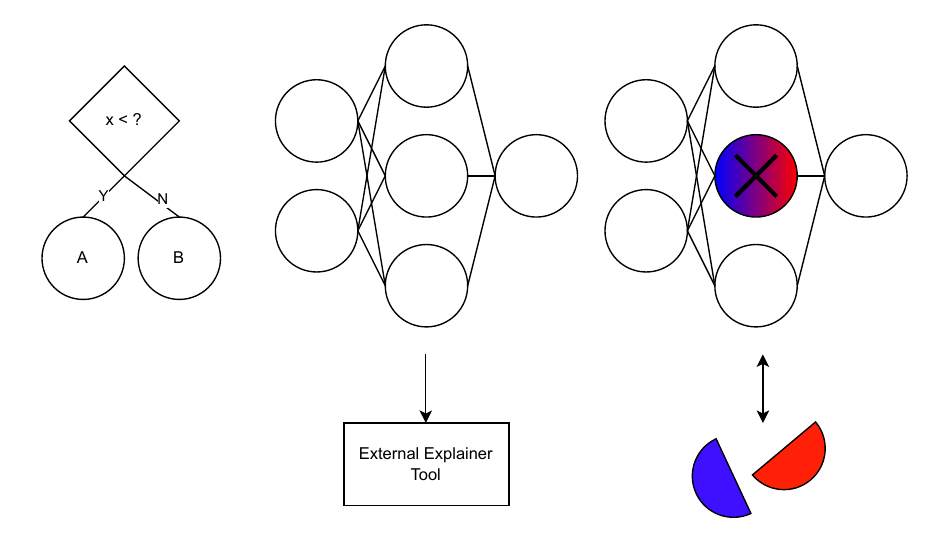}
    \caption{Interpretability approaches. Left: Inherent (transparent by design). Center: Post-hoc (external explainer tools). Right: Mechanistic (reverse-engineering through ablation).}
    \label{fig:approach}
\end{figure}

The third approach, mechanistic interpretability, investigates how models solve  classes of problems,  investigating the ``mechanisms underlying neural network generalization''~\citep{sharkey2025openproblems}. Rather than explaining individual predictions or building transparent architectures, this approach reverse-engineers learned algorithms~(Figure~\ref{fig:approach}).

\subsection{Transformer Architecture Background}

We briefly review the core transformer components most relevant to mechanistic interpretability analysis, focusing on attention mechanisms. For comprehensive architectural details, see \citet{vaswani2017attention}.

\subsubsection{Self-Attention Mechanism}
The self-attention mechanism transforms input sequences using queries, keys, and values. Given input matrices where queries $Q$ and keys $K$ have dimension $d_k$, and values $V$ have dimension $d_v$, the scaled dot-product attention computes:

\begin{align}
\text{Attention}(Q,K,V) = \text{softmax}\left(\frac{QK^\top}{\sqrt{d_k}}\right)V
\end{align}

The input dimensions are $Q, K \in \mathbb{R}^{n \times d_k}$ and $V \in \mathbb{R}^{n \times d_v}$, where $n$ represents the sequence length. The scaling factor $\frac{1}{\sqrt{d_k}}$ prevents gradient vanishing in the softmax function.

\subsubsection{Multi-Head Attention}

Multi-head attention computes attention in parallel across $h$ independent heads, each operating on a lower-dimensional subspace:

\begin{align}
\text{MultiHead}(Q, K, V) = \text{Concat}(\text{head}_1, \ldots, \text{head}_h)W^O 
\end{align}

where 

\begin{align}
\text{head}_i = \text{Attention}(QW^Q_i, KW^K_i, VW^V_i) 
\end{align}

The projection matrices $W^Q_i, W^K_i, W^V_i$ have dimension $d_{\text{model}} \times d_k$ where $d_k = d_{\text{model}}/h$, and the output projection $W^O \in \mathbb{R}^{d_{\text{model}} \times d_{\text{model}}}$ combines the concatenated outputs.

This parallel, modular structure is important for mechanistic interpretability. Each head computes attention independently, enabling targeted interventions where individual heads can be selectively ablated while others remain functional. This architectural modularity makes systematic causal analysis possible.

\subsection{Why Attention Seemed Interpretable}

Before examining why attention seemed interpretable, it is important to situate attention weight analysis within our interpretability taxonomy. Attention visualization and weight-based analysis fall within the post-hoc explainability paradigm rather than mechanistic interpretability proper. These methods observe correlations in trained models but do not reverse-engineer the underlying computational mechanisms through causal intervention. While useful for generating hypotheses, they require validation through mechanistic methods like ablation to establish genuine causal understanding.

The attention mechanism's interpretability potential was recognized early. Unlike 
opaque recurrent states, attention weights form explicit token-to-token maps that 
appear similar to human reading patterns. This transparency motivated extensive 
research into whether attention patterns provide faithful explanations of model 
behaviour.

Initial studies visualized attention patterns and found apparent linguistic 
structures, suggesting these patterns might reveal how models process language 
\citep{clark2019does}. Systematic analyses discovered heads specialized for syntactic 
dependencies. These findings reinforced the intuition 
that attention heads encode interpretable linguistic functions.

\subsection{The Need for Causal Evidence}

Do attention patterns reliably indicate what models 
use to make decisions? Early work assumed attention weights revealed causal 
importance. If a model attends strongly to certain tokens, those tokens should 
matter for decisions. This assumption was rarely stated explicitly, but attention 
visualizations became ubiquitous, with patterns treated as explanations of model 
behaviour.

However, this assumption came under sustained empirical scrutiny. Multiple studies 
demonstrated that attention patterns could be altered dramatically without changing 
predictions, and that tokens receiving high attention were not always causally 
important \citep{jain2019attention, serrano2019attention}. These findings suggested 
that observing attention patterns provides insufficient evidence for causal 
importance. This motivated a shift from observational methods (visualization) to 
interventional methods (ablation).

\subsection{The Focus of This Paper}

This paper traces how key papers established the current framework. The discussion centers on natural language processing, where the majority of mechanistic interpretability research has been conducted, though the principles and methods generalize to any domain.
\citet{serrano2019attention} demonstrated that attention patterns don't reliably 
indicate causal importance. \citet{wiegreffe2019attention} resolved this by 
distinguishing plausibility from faithfulness, arguing causal intervention is 
necessary. \citet{jacovi2020towards} formalized faithfulness into testable 
criteria. \citet{michel2019sixteen} applied these ideas through systematic head 
ablation, revealing both specialization and massive redundancy.

Head intervention studies have revealed a fundamental tension between specialization 
and redundancy. Individual attention heads develop specialized, interpretable 
functions for specific linguistic patterns. Yet models exhibit substantial 
redundancy, with a majority of heads removable without catastrophic performance 
loss \citep{michel2019sixteen, voita2019analyzing}. These findings suggest 
attention mechanisms balance specialization with robustness.

We argue that head intervention provides stronger evidence than visualization or 
correlation-based methods for understanding attention mechanisms. We examine recent work demonstrating that identified attention heads can be actively manipulated to control model behaviour, validating the practical utility of head-level interpretability. However, 
methodological challenges remain, including distribution shift problems and 
polysemanticity. Despite these limitations, causal head intervention has enabled practical 
compression techniques and reshaped how researchers approach transformer 
interpretability.

\section{The Attention Explanation Debate}
The interpretation of attention mechanisms evolved from treating patterns as direct explanations to recognizing the necessity of causal intervention.  It is important to note that attention visualization and correlation-based analysis of attention weights fall primarily within the post-hoc explainability paradigm rather than mechanistic interpretability. While these methods can generate plausible hypotheses about model behaviour, they do not reverse-engineer the underlying computational mechanisms. True mechanistic interpretability requires causal intervention to establish which components actually drive model computations. The debate 
proceeded in four phases: Early visualization studies that assumed attention reveals 
importance, critiques demonstrating this assumption fails, a conceptual resolution 
distinguishing plausibility from faithfulness, and formalization of testable criteria 
that motivated head intervention.

\subsection{Early Attention Visualization Studies}

Initial transformer interpretability research treated attention weights as direct explanations of model decisions. Attention appeared interpretable because its patterns were visually intuitive and seemed to encode linguistic structure. Visualization tools \citep{vig2019multiscale} plotted which input tokens each attention head focused on, creating maps of the model's internal focus. \cite{clark2019does} demonstrated that specific heads specialized in distinct syntactic roles, such as linking verbs to direct objects (e.g., ``baked'' to ``cake''), resolving coreference (e.g., ``her'' to ``lawyer''), or forming determiner-noun phrases. These clear patterns fueled the hypothesis that attention weights could explain model behaviour

\subsection{Challenges to Attention as Explanation}

The assumption that attention provides faithful explanations faced mounting challenges. 
Multiple studies demonstrated fundamental limitations of analysing attention through 
visualization alone, ultimately motivating the shift toward causal head intervention 
methods.

\subsubsection{Different Attention Patterns Can Produce Identical Outputs}

\citet{jain2019attention} challenged the explanatory value of attention weights 
through experiments demonstrating that alternative attention distributions 
(dramatically different from original learned patterns) could produce identical 
predictions. These results established that multiple attention patterns can 
implement the same computation. Observing one particular pattern does not prove 
it is necessary or that it represents the only valid explanation. This non-uniqueness 
problem suggested that visualization alone cannot determine which patterns matter 
causally.

\subsubsection{Weak Correlation Between Attention and Importance}

\citet{serrano2019attention} conducted systematic experiments comparing attention 
weights to gradient-based importance measures. Across multiple architectures and 
tasks, they found weak correlation between high-attention tokens and those identified 
as most important by gradient measures. Removing high-attention tokens often had 
minimal impact, while removing low-attention tokens sometimes caused catastrophic 
failures.

The weak correlation between attention weights and various importance measures 
suggested attention patterns might be task-specific artifacts rather than reliable 
indicators of causal structure. This finding directly motivates ablation. If we
can not trust observational measures to reveal causal importance, we need direct 
intervention to test what components actually do when removed.

Together, these theoretical and empirical results converged on a conclusion. Raw 
attention weights provide insufficient evidence for causal importance. Visualization 
reveals patterns but cannot distinguish whether those patterns drive behaviour or 
merely correlate with it.

\subsubsection{High Attention Can Have Little Functional Impact}

Early ablation studies revealed surprising disconnects between attention magnitude and functional importance. \citet{michel2019sixteen} showed that heads with highest average attention weights were not necessarily most important for task performance. Some prominent heads could be removed with negligible impact, while seemingly minor heads proved critical. \citet{voita2019analyzing} found that positional heads attending to adjacent tokens with high weights were among the least important by ablation metrics. While rare-word heads with lower average attention proved essential for handling unusual inputs. This inversion of expected importance highlighted the necessity of causal rather than correlational analysis.

\section{Plausibility versus Faithfulness}

These critiques created apparent tension: if attention patterns are neither unique 
nor correlated with importance, what interpretive value do they have? 
\citet{wiegreffe2019attention} resolved this apparent contradiction. Their key 
contribution was distinguishing two separate properties that explanations might 
possess.

Plausibility measures whether explanations align with human intuitions. Do they make sense to domain experts examining the explanation? An attention pattern showing high weights on semantically relevant tokens is plausible. It matches what humans expect important tokens to look like. If a sentiment classifier attends to emotionally charged words when predicting sentiment, this seems reasonable. This holds regardless of whether those tokens are causally necessary. Plausibility serves important purposes. It enables humans to build mental models of system behaviour. It facilitates trust in deployed systems. It allows domain experts to apply their knowledge to understanding model decisions~\citep{kadem2025human}.

Faithfulness measures something different. Whether explanations accurately represent the model's actual decision-making process. Whether they correctly predict how interventions affect model behaviour. An explanation is faithful if removing components it identifies as important degrades performance proportionally. Removing components it identifies as unimportant has negligible effect. Faithfulness concerns causal accuracy. Does the explanation correctly identify what the model actually uses?

The important insight is that these properties are independent. Attention patterns can be plausible without being faithful. They reveal patterns that match human intuitions about language understanding. But they may not accurately represent causal structure. The model might produce sensible-looking attention distributions that happen to correlate with behaviour. Those distributions might not causally drive the behaviour.

Conversely, faithful explanations need not be plausible. The model's true mechanisms 
may differ fundamentally from human problem-solving strategies. Models may implement 
valid solutions using algorithms unintuitive to humans, yet these remain faithful 
if they accurately predict behaviour under intervention.

This resolution revealed that observational analysis provides correlational information. Examining activation patterns or attention weights without intervention cannot establish causality. We can observe what patterns emerge. But observation alone cannot establish whether those patterns causally drive behaviour or merely correlate with it.

\section{The Formalization}

While \citet{wiegreffe2019attention} distinguished plausibility from faithfulness 
conceptually, \citet{jacovi2020towards} formalized this distinction into testable 
criteria. They defined faithfulness operationally through three properties. 
Comprehensiveness requires that removing components identified as important 
should substantially degrade performance. Sufficiency requires that 
removing components identified as unimportant should not substantially affect 
performance. Invariance requires that explanations should not change for 
functionally equivalent model variants.

These properties operationalize the plausibility-faithfulness distinction. 
Plausibility concerned whether explanations match human intuitions. Faithfulness 
concerned whether they predict intervention effects. \citet{jacovi2020towards} 
made faithfulness testable: explanations are faithful if they satisfy 
comprehensiveness (correctly predict what matters) and sufficiency (correctly 
predict what does not matter). This transformation from philosophical distinction 
to empirical test enabled systematic evaluation of interpretability methods.

However, their framework rests on assumptions worth noting. \citet{jacovi2020towards} 
assume component independence: removing component $A$ has the same effect regardless 
of component $B$'s presence. This assumption frequently fails in neural networks. 
Attention heads interact through the residual stream. Layer normalisation couples 
representations. Downstream components compensate for ablated components.

When independence fails, comprehensiveness and sufficiency become ambiguous. If 
removing $A$ has different effects depending on $B$'s presence, importance is not 
an inherent property but emerges from interactions. The framework also struggles 
with redundancy: if components $A$ and $B$ implement the same function redundantly, 
removing either individually has small effect (suggesting unimportance) but removing 
both has large effect (suggesting importance).

Despite these limitations, the framework provided crucial conceptual clarity. It 
established that causal claims require causal evidence. This motivated interventional 
methods, particularly head ablation.

\section{Resolution}

The resolution emerged that attention provides partial but valuable explanatory 
signal when properly tested through causal intervention. Neither wholesale dismissal 
nor uncritical acceptance is warranted. Causal intervention methods, particularly 
head ablation, became essential for distinguishing correlation from causation in 
attention analysis.

Attention visualization remains useful for generating hypotheses and providing 
intuition about model behaviour. However, establishing which patterns actually 
drive behaviour requires causal testing. This motivated the development of head 
ablation methodologies discussed in the next section.

\section{Head Ablation Methodology and Key Findings}

\subsection{What is Attention Head Ablation?}

Attention head ablation tests the causal importance of individual attention heads through a simple intervention. Remove a head, measure the performance change, and infer its necessity for task behaviour \citep{michel2019sixteen}. Unlike visualization methods that observe existing patterns, head ablation provides causal evidence by actively manipulating the model.

The fundamental principle involves comparing model behaviour with and without specific heads, quantifying their contribution to task performance. This approach establishes which heads genuinely drive model behaviour rather than merely correlating with outputs, directly addressing the faithfulness criterion established by \citet{jacovi2020towards}.

\subsection{Head Ablation Methodology}

\subsubsection{Zero Ablation}

Head ablation removes individual attention heads by setting their outputs to zero. \citet{michel2019sixteen} introduced this method, testing each head's importance by measuring accuracy drops when the head is removed. Zero ablation completely eliminates the head's contribution, providing a clear test of necessity.

For a model with $L$ layers and $H$ heads per layer, this requires $L \times H$ separate evaluations to test each head individually. \citet{michel2019sixteen} tested BERT's 144 heads across multiple tasks, finding that most heads contributed minimally to performance. This addresses the concerns raised by \citet{serrano2019attention} about whether attention patterns reflect genuine importance.

\subsubsection{Mean Ablation}

Recent work explores alternatives to zero ablation. Mean ablation replaces removed heads with average values computed over the dataset, reducing the risk of producing unusual activations that never occur during normal operation. This partially addresses distribution shift concerns discussed in Chapter 3.

However, the core principle remains the same across variants. Remove the head and measure what changes. The key insight is causal intervention rather than observation.

\subsubsection{Learned Pruning}

\citet{voita2019analyzing} refined head ablation by learning which heads are important during training rather than testing each head individually after training. Their method introduces learnable binary masks for each head, using regularization to encourage the model to rely on fewer heads. This automatically identifies which heads are essential.

The advantage of learned pruning is efficiency. Rather than testing hundreds of head combinations, the model learns an optimal sparse configuration during training. The disadvantage is that importance scores depend on the specific regularization strength and optimization procedure.

\section{Specialization of Attention Heads}

Head ablation studies conclusively demonstrate that attention heads specialize for different functions in language models. \citet{olsson2022context} identified induction heads implementing in-context learning through systematic ablation. These heads compose information from previous occurrences of patterns to predict repeated sequences. Removing induction heads selectively impairs few-shot learning while preserving other capabilities.

Other work discovered heads implementing indirect object identification in GPT-2, 
tracing the complete computational circuit through ablation 
\citep{wang2023interpretability}. Specific heads in middle layers proved necessary 
for this task, while other heads contributed minimally. This kind of selective 
impairment provides strong evidence for functional specialization, supporting 
\citet{wiegreffe2019attention}'s argument that attention can be explanatory when 
tested causally.


The specialization finding means different heads serve different purposes. Some focus on syntactic relationships like subject-verb agreement, others on semantic content, and still others on positional information. This division of labor emerges during training without explicit supervision.

\section{Redundancy}

A consistent finding across head ablation studies is substantial redundancy in 
attention mechanisms. \citet{michel2019sixteen} showed 70-90\% of attention heads 
in BERT are removable with minimal performance loss. \citet{voita2019analyzing} 
achieved 92\% of original performance on machine translation while retaining only 
17\% of heads. This redundancy is not uniform across heads. Studies found 
importance follows a power-law distribution. A few heads are critical while most 
are dispensable.

\citet{voita2019analyzing} demonstrated this application in neural machine 
translation. They systematically ablated each of 48 attention heads in a 6-layer 
transformer, measuring BLEU score changes. Results showed 17 heads caused 
substantial drops. Another 13 caused moderate drops. The remaining 18 caused 
minimal drops.

Importantly, heads with large ablation effects did not always have the most 
interpretable attention patterns. Some heads with clear, human-interpretable 
patterns were actually dispensable. Meanwhile, some heads with complex, less 
interpretable patterns were critical. This validated the debate's conclusion: 
plausibility (interpretable patterns) does not guarantee faithfulness (causal 
importance).

They then iteratively pruned heads with smallest effects. After each removal, they 
re-evaluated remaining heads to account for potential compensation. The process 
successfully removed 79\% of the model while translation quality stayed within 
0.15 BLEU of the original. The pruned model enabled substantially faster inference, 
requiring only 21\% of attention parameters. This demonstrates that architectural ablation suffices for compression. Researchers 
did not need to understand what computations each head performed internally. They 
only needed to know which architectural components were removable while maintaining 
acceptable performance.

The redundancy finding appears functional rather than 
wasteful. It provides robustness, allowing models to maintain performance even 
when individual components fail or are removed. This has important implications 
for the interpretability debate. Even if many heads are removable, the remaining 
critical heads may still provide faithful explanations of model behaviour 
\citep{wiegreffe2019attention}.

\section{Hierarchical Organization}

Head ablation reveals hierarchical organization within attention layers. Early heads compute primitive features that later heads compose into complex representations. Ablating early heads causes cascading failures in dependent later heads, while late head ablation has more localized effects. This suggests a compositional structure where simple patterns are detected first and combined progressively.

\citet{clark2019does} traced how syntactic structure emerges hierarchically 
in BERT. They found Layer 2-4 heads detect basic grammatical relationships 
(noun-verb, determiner-noun). Layer 5-8 heads compose these into longer-range 
dependencies (subject-verb across clauses). Layer 9-12 heads integrate 
sentence-level structure. Ablation confirmed this hierarchy: removing early 
heads degraded both basic and complex syntax, while removing late heads 
impaired only complex dependencies. Validation through structured ablation confirmed this hierarchical organization.

\section{Negative Interactions Between Heads}

Head ablation revealed unexpected negative interactions between components. Some studies discovered cases where removing certain heads actually improves performance. These effects occur when harmful heads introduce interference or noise that degrades overall computation. \citet{wang2023interpretability} identified suppression heads in GPT-2 
that inhibit indirect object retrieval. Layer 9 Head 9 actively suppressed  predictions by writing negative contributions to the residual stream. 
Ablating this head improved accuracy by 12 percentage points on targeted 
inputs, revealing latent capability that normal model operation blocked.

These findings complicate the interpretation of redundancy. Some heads may appear redundant because they are being suppressed by other heads. Measuring importance requires considering not just individual heads but their interactions. This nuance is important for the kind of faithful interpretation advocated by \citet{jacovi2020towards}.

\section{Comparison with Other Methods}

Head ablation differs from other interpretability approaches in providing causal rather than correlational evidence. Attention visualization methods \citep{vig2019multiscale, clark2019does} display attention patterns and generate hypotheses about model behaviour. However, as \citet{serrano2019attention} demonstrated, high attention weights do not guarantee causal importance. Visualization is useful for exploration but requires ablation for validation.

Other methods include attention rollout, which recursively multiplies attention matrices to track information flow \citep{abnar2020quantifying}, and various attribution techniques that identify important input features. These methods offer computational efficiency but cannot establish the causal relationships that head ablation provides. \citet{serrano2019attention} found gradient-based importance scores 
correlated only 0.28 (Spearman) with ablation-based rankings across BERT 
heads. Attention rollout \citep{abnar2020quantifying} achieved 0.35 
correlation. Integrated gradients performed slightly better at 0.41. 
All methods showed substantial disagreement about which heads matter most, 
with top-10 rankings overlapping by only 3-4 heads on average.

\citet{jacovi2020towards} established that head ablation uniquely satisfies faithfulness criteria because it tests causal necessity through intervention. Other methods may identify correlations or surface patterns, but only ablation directly tests what happens when components are removed. This makes head ablation the strongest available method for validating mechanistic explanations, though it can be combined with other approaches for comprehensive analysis.

\section{Why Head Ablation Provides Faithful Explanations}

\citet{jacovi2020towards} argued that ablation-based explanations satisfy 
faithfulness criteria that correlation-based methods fail. By establishing causal 
rather than correlational relationships, head ablation provides stronger causal evidence 
about component importance. If removing a head degrades performance, that head must 
have been contributing causally. This directly addresses the concerns raised by 
\citet{serrano2019attention} about whether attention patterns genuinely drive model 
behaviour.

The faithfulness framework translates directly to ablation. Comprehensiveness 
requires that components identified as important should cause substantial 
performance drops when removed. Sufficiency requires that components identified 
as unimportant should cause negligible performance drops. Together, these properties 
enable testing whether importance attributions predict intervention effects 
accurately.

\citet{wiegreffe2019attention} demonstrated that head ablation can validate 
attention's explanatory value in ways that correlation analysis cannot. When 
evaluated through ablation, attention patterns often prove causally important, 
even though multiple attention patterns could theoretically produce similar outputs. 
The key insight is that the existence of alternative patterns does not invalidate 
the causal importance of the observed pattern.

However, head ablation's faithfulness comes with caveats discussed in the next 
section. Distribution shift and polysemanticity can complicate interpretation. 
Despite these limitations, head ablation remains the strongest available method 
for establishing causal relationships in neural networks, making it essential for 
testing the kind of faithfulness that \citet{jacovi2020towards} defined.

\section{From Interpretation to Control}

The preceding sections established that attention head ablation reveals which heads 
causally contribute to model behaviour. A natural question follows. Can we leverage 
this understanding to \textit{control} model outputs? Recent work demonstrates that 
we can, validating the practical utility of mechanistic interpretability.

\citet{basile2025headpursuit} introduced a method for identifying and manipulating 
attention heads specialized for semantic properties in both language and 
vision-language models. Their approach uses Simultaneous Orthogonal Matching Pursuit 
(SOMP) to discover which vocabulary directions best explain each head's outputs, 
effectively labeling heads by their semantic function. The method reveals specificity. Individual heads in Mistral-7B 
specialize for politics, nationality, temporal concepts, and numerical reasoning, 
with each head's top-attended tokens clustering coherently around its semantic category.

Critically, the method moves beyond observation to intervention. By rescaling 
the residual stream contributions of identified heads, model outputs can be 
systematically enhanced or suppressed along semantic dimensions. Setting a 
negative scaling factor ($\alpha = -1$) on color-specialized heads causes a 
vision-language model to omit color descriptions entirely; amplification 
($\alpha = 5$) produces outputs saturated with color terms. Similar control 
extends to sentiment, quantity, and abstract categories.

The intervention results provide strong validation of head-level interpretability. 
On question-answering tasks, ablating heads identified by SOMP for 
``country''-related content selectively degrades performance on geography 
questions while preserving accuracy on other categories. This selective impairment demonstrates faithful mechanistic explanation as defined by \cite{jacovi2020towards}, confirming that the identified specializations reflect genuine causal structure.

Perhaps most significantly for safety applications, the approach enables 
toxicity reduction without retraining. Suppressing 32 heads identified as 
toxicity-related reduced toxic generations by 34--51\% across benchmarks, 
substantially outperforming random head selection. This result suggests that 
undesirable behaviors may be localized to specific attention heads amenable 
to targeted intervention.

These findings represent a maturation of the field from interpretation to 
control. Where early attention visualization generated hypotheses and ablation 
tested necessity, targeted intervention now enables behavioural modification 
grounded in mechanistic understanding. The progression validates the core 
premise of mechanistic interpretability. Understanding \textit{how} 
models compute enables principled modification of \textit{what} they compute.

\section{Limitations and Open Problems}

\subsection{The Distribution Shift Problem}

Head ablation fundamentally alters model activations, potentially creating unusual 
states that never occur during normal operation. Research demonstrated that zero 
ablation creates activations outside the training distribution, triggering 
unpredictable behaviour unrelated to head importance \citep{hase2021out}. Studies 
showed that distribution shift effects scale with model depth and width. Deeper 
models exhibit greater sensitivity, with late-layer ablations causing cascading 
distribution violations.

\citet{hase2021out} demonstrated that ablation creates out-of-distribution 
activations, with magnitude shifts up to 4-5× normal values in deep models. This 
distribution shift means observed failures may reflect artificial model states 
rather than genuine importance, complicating causal interpretation. Ablation must intervene strongly enough to 
measure effects but not so strongly that it breaks the model in artificial ways. 
Mean ablation partially addresses this by replacing removed heads with typical 
values, but still introduces distribution shift for unusual inputs.

Current solutions remain imperfect. More sophisticated ablation methods reduce 
distribution shift but increase computational cost or sacrifice the clean causal 
interpretation of simple ablation. The field lacks principled methods for 
quantifying when distribution shift invalidates conclusions. This is a key 
limitation for the faithful interpretability framework proposed by 
\citet{jacovi2020towards}.

\subsection{Polysemanticity}

Polysemanticity poses a challenge for attention interpretation. Neural networks represent far more features than they have dimensions~\citep{elhage2022toy}, and this phenomenon extends to attention. A single attention head in a four-layer transformer simultaneously serves multiple functions depending on context~\citep{janiak2023polysemantic}. If attention heads exhibit superposition, ablating a single head disrupts multiple unrelated functions simultaneously, making it difficult to isolate specific mechanisms through single-head interventions alone.

This directly challenges the goal of faithful interpretation advocated by 
\citet{jacovi2020towards}. If we cannot isolate what a head does, ablation tells 
us the head is important but not what function it serves. Addressing 
polysemanticity requires methods that can isolate distinct functions within 
individual heads, though such decomposition remains an open research challenge. Polysemanticity appears to increase with model scale. Larger models pack more 
functionality into individual heads, making the problem worse precisely where 
interpretability matters most for safety and alignment.

\subsection{Measurement and Scalability Challenges}

Head ablation conclusions depend critically on how performance is measured. 
Research showed that different metrics (accuracy, perplexity, loss) yield 
contradictory importance rankings for the same attention heads 
\citep{michel2019sixteen}. Studies found importance rankings vary significantly 
across different datasets and tasks. Heads essential for one distribution prove 
redundant for another, questioning the generalizability of ablation findings.

Exhaustive head ablation becomes computationally prohibitive for large language 
models. Testing each attention head individually in models with thousands of heads 
requires thousands of forward passes. Even with efficient approximations, 
identifying all important heads in frontier models exceeds practical computational 
budgets.

For practical research, this means head ablation studies often focus on smaller 
models like BERT-base or GPT-2, where exhaustive analysis remains feasible. 
Findings from these models may not fully generalise to much larger models.

\subsection{Causal Interpretation}

Even with causal interventions, interpretation requires care. Research demonstrated 
backup heads that activate only under ablation, creating apparent redundancy where 
none exists during normal operation \citep{voita2019analyzing}. These artifacts 
mean observed robustness may not reflect how the model actually operates in 
deployment.

Current ablation methods primarily test necessity (whether components are required) 
rather than sufficiency (whether components fully explain behaviour). This 
limitation means ablation establishes that heads contribute causally without 
proving they are the sole or complete explanation.

Understanding these subtleties is important for avoiding overconfident conclusions. 
Head ablation provides strong evidence about component importance but does not 
reveal the complete picture of how heads interact and compensate for each other. 
This is relevant to \citet{wiegreffe2019attention}'s point that multiple valid 
explanations can coexist.

\subsection{Cross-Modal Attention Mechanisms}

Current head ablation research focuses almost exclusively on self-attention. Cross-attention mechanisms 
in multimodal models remain largely unexplored. Unlike 
self-attention, where queries and keys come from the same sequence, cross-attention 
attends from one modality or sequence to another. This creates
different interpretability challenges.

The challenge is confounded dependencies. Ablating encoder heads affects 
cross-attention inputs. Ablating decoder self-attention affects cross-attention 
queries. Ablating cross-attention affects decoder states fed to subsequent layers. 
Clean causal attribution requires careful experimental design that separates these 
effects.

Multimodal systems create additional complexity. Transformers are increasingly 
used for fusing heterogeneous data, combining sensor modalities, These systems use cross-attention between 
different data types. Does cross-modal ablation behave like within-modality 
ablation? Or do modality-bridging mechanisms exhibit different redundancy patterns?

Understanding cross-attention is particularly important for safety-critical 
applications. If certain heads are essential for integrating complementary 
information sources under adverse conditions, their ablation would cause failures 
invisible in single-modality testing. Developing ablation methodology for 
cross-attention would expand mechanistic interpretability beyond single-modality 
models and enable testing hypotheses about how models align and integrate 
information across representations.

\subsection{Other Future Directions}

Despite these limitations, several promising directions exist for future research. 
Developing intervention methods that intervene without pushing models into unrealistic 
states is key to ensuring validity of causal claims. Integrating ablation with 
methods that can decompose polysemantic heads would enable fine-grained mechanistic 
understanding while maintaining causal rigour.

Developing scalable yet faithful head ablation techniques remains a central 
challenge. Future methods must balance computational efficiency with causal 
validity. This likely requires fundamental advances beyond current approaches, 
potentially involving smarter search strategies or theoretical guarantees about 
when exhaustive testing is unnecessary.

Extending ablation methodology to cross-attention mechanisms represents a natural 
next step as transformers expand into multimodal and multi-sequence architectures. 
Understanding how cross-attention differs from self-attention could reveal 
important architectural principles for building more interpretable systems.

\section{Conclusion}

Head intervention studies revealed three robust empirical findings. First, attention heads exhibit both specialization and massive redundancy. Individual heads develop distinct functions for specific linguistic patterns, yet 70 to 90 percent remain removable without catastrophic failure. This paradox suggests transformers balance specialization with robustness through redundant pathways. Second, plausible attention patterns do not guarantee causal importance. Heads with interpretable linguistic patterns often prove dispensable, while critical heads may implement opaque computations. This finding validates the distinction between plausibility and faithfulness in neural network explanations. Third, heads interact through complex dynamics including suppression and negative contributions, not merely additive combination of features.

Recent work has validated that mechanistic understanding enables targeted behavioural control. Identifying heads specialized for semantic properties allows systematic manipulation of model outputs through attention rescaling. This progression from interpretation to intervention represents a maturation of the field. Where early visualization generated hypotheses and ablation tested necessity, researchers can now modify model behaviour based on mechanistic understanding. The ability to reduce toxic outputs through targeted head suppression demonstrates that interpretability research has practical safety applications, moving beyond purely scientific investigation toward enabling safer AI systems.

These findings have enabled practical applications in model compression. Ablation-guided pruning removes most attention heads while maintaining performance within acceptable bounds. More fundamentally, head intervention provides the first rigorous methodology for testing mechanistic hypotheses about transformer internals. Researchers can now validate causal claims rather than relying on correlational evidence.

Significant limitations constrain current ablation methods. Distribution shift creates artificial activation patterns that may not reflect normal model operation. Polysemanticity prevents clean isolation of individual functions within heads. Scalability constraints limit exhaustive analysis to smaller models. These challenges indicate that full mechanistic understanding requires methodological advances beyond current approaches.
Head ablation established a critical principle for interpretability research. Causal intervention is necessary to distinguish correlation from causation in neural networks. This standard now shapes mechanistic interpretability across architectures and domains. The methods developed for attention analysis provide templates for investigating other neural network components.
The shift from attention visualization to head ablation demonstrates that understanding neural networks requires the same scientific rigor applied in other empirical fields. Observation alone cannot establish mechanism. Intervention reveals function. This methodological advance has helped transform interpretability from exploratory analysis into hypothesis-driven science grounded in causal evidence.

\bibliographystyle{unsrtnat}
\bibliography{custom}

@inproceedings{basile2025headpursuit,
  title={Head Pursuit: Probing Attention Specialization in Multimodal Transformers},
  author={Basile, Lorenzo and Maiorca, Valentino and Doimo, Diego and Locatello, Francesco and Cazzaniga, Alberto},
  booktitle={Advances in Neural Information Processing Systems},
  year={2025}
}

@article{vaswani2017attention,
  title={Attention is all you need},
  author={Vaswani, Ashish and Shazeer, Noam and Parmar, Niki and Uszkoreit, Jakob and Jones, Llion and Gomez, Aidan N and Kaiser, {\L}ukasz and Polosukhin, Illia},
  journal={Advances in Neural Information Processing Systems},
  volume={30},
  year={2017}
}

@inproceedings{jacovi2020towards,
  title={Towards faithfully interpretable NLP systems: How should we define and evaluate faithfulness?},
  author={Jacovi, Alon and Goldberg, Yoav},
  booktitle={Proceedings of the 58th Annual Meeting of the Association for Computational Linguistics},
  pages={4198--4205},
  year={2020}
}

@inproceedings{serrano2019attention,
  title={Is attention interpretable?},
  author={Serrano, Sofia and Smith, Noah A},
  booktitle={Proceedings of the 57th Annual Meeting of the Association for Computational Linguistics},
  pages={2931--2951},
  year={2019}
}

@inproceedings{wiegreffe2019attention,
  title={Attention is not not explanation},
  author={Wiegreffe, Sarah and Pinter, Yuval},
  booktitle={Proceedings of the 2019 Conference on Empirical Methods in Natural Language Processing},
  pages={11--20},
  year={2019}
}

@inproceedings{clark2019does,
  title={What does BERT look at? An analysis of BERT's attention},
  author={Clark, Kevin and Khandelwal, Urvashi and Levy, Omer and Manning, Christopher D},
  booktitle={Proceedings of the 2019 ACL Workshop BlackboxNLP},
  pages={276--286},
  year={2019}
}

@article{michel2019sixteen,
  title={Are sixteen heads really better than one?},
  author={Michel, Paul and Levy, Omer and Neubig, Graham},
  journal={Advances in Neural Information Processing Systems},
  volume={32},
  year={2019}
}

@inproceedings{voita2019analyzing,
  title={Analyzing multi-head self-attention: Specialized heads do the heavy lifting, the rest can be pruned},
  author={Voita, Elena and Talbot, David and Moiseev, Fedor and Sennrich, Rico and Titov, Ivan},
  booktitle={Proceedings of the 57th Annual Meeting of the Association for Computational Linguistics},
  pages={5797--5808},
  year={2019}
}

@inproceedings{jain2019attention,
  title={Attention is not explanation},
  author={Jain, Sarthak and Wallace, Byron C},
  booktitle={Proceedings of the 2019 Conference of the North American Chapter of the Association for Computational Linguistics},
  pages={3543--3556},
  year={2019}
}

@article{olsson2022context,
  title={In-context learning and induction heads},
  author={Olsson, Catherine and Elhage, Nelson and Nanda, Neel and Joseph, Nicholas and DasSarma, Nova and Henighan, Tom and Mann, Ben and Askell, Amanda and Bai, Yuntao and Chen, Anna and others},
  journal={Transformer Circuits Thread},
  year={2022}
}

@inproceedings{wang2023interpretability,
  title={Interpretability in the wild: A circuit for indirect object identification in GPT-2 small},
  author={Wang, Kevin and Variengien, Alexandre and Conmy, Arthur and Shlegeris, Buck and Steinhardt, Jacob},
  booktitle={International Conference on Learning Representations},
  year={2023}
}

@inproceedings{vig2019multiscale,
  title={A multiscale visualization of attention in the transformer model},
  author={Vig, Jesse},
  booktitle={Proceedings of the 57th Annual Meeting of the Association for Computational Linguistics: System Demonstrations},
  pages={37--42},
  year={2019}
}

@inproceedings{abnar2020quantifying,
  title={Quantifying attention flow in transformers},
  author={Abnar, Samira and Zuidema, Willem},
  booktitle={Proceedings of the 58th Annual Meeting of the Association for Computational Linguistics},
  pages={4190--4197},
  year={2020}
}

@article{elhage2022toy,
  title={Toy models of superposition},
  author={Elhage, Nelson and Hume, Tristan and Olsson, Catherine and Schiefer, Nicholas and Henighan, Tom and Kravec, Shauna and Hatfield-Dodds, Zac and Lasenby, Robert and Drain, Dawn and Chen, Carol and others},
  journal={arXiv preprint arXiv:2209.10652},
  year={2022}
}

@article{janiak2023polysemantic,
  title={Polysemantic Attention Head in a 4-Layer Transformer},
  author={Janiak, Jett and Mathwin, Chris and Hex, Stefan},
  journal={AI Alignment Forum},
  year={2023},
  url={https://www.alignmentforum.org/posts/nuJFTS5iiJKT5G5yh/polysemantic-attention-head-in-a-4-layer-transformer}
}

@article{hase2021out,
  title={The out-of-distribution problem in explainability and search methods for feature importance explanations},
  author={Hase, Peter and Bansal, Mohit},
  journal={Advances in Neural Information Processing Systems},
  volume={34},
  pages={3650--3666},
  year={2021}
}

@article{sharkey2025openproblems,
  title={Open Problems in Mechanistic Interpretability},
  author={Sharkey, Lee and Chughtai, Bilal and Batson, Joshua and Lindsey, Jack and Wu, Jeff and Bushnaq, Lucius and Goldowsky-Dill, Nicholas and Heimersheim, Stefan and Ortega, Alejandro and Bloom, Joseph and Biderman, Stella and Garriga-Alonso, Adri{\`a} and Conmy, Arthur and Nanda, Neel and Rumbelow, Jessica and Wattenberg, Martin and Schoots, Nandi and Miller, Joseph and Michaud, Eric J. and Casper, Stephen and Tegmark, Max and Saunders, William and Bau, David and Todd, Eric and Geiger, Atticus and Geva, Mor and Hoogland, Jesse and Murfet, Daniel and McGrath, Thomas},
  journal={arXiv preprint arXiv:2501.16496},
  year={2025}
}

@inproceedings{kadem2023xgboost,
  title={XGBoost for Interpretable Alzheimer's Decision Support},
  author={Kadem, Mason and Noseworthy, Michael and Doyle, Thomas},
  booktitle={Proceedings of the AAAI Symposium Series},
  volume={1},
  number={1},
  pages={135--141},
  year={2023}
}

@mastersthesis{kadem2023interpretable,
  title={Interpretable Machine Learning in Alzheimer's Disease Dementia},
  author={Kadem, Mason},
  year={2023},
  school={McMaster University}
}

@inproceedings{kadem2025sleep,
  author    = {Kadem, Mason and Zheng, Rong},
  title     = {Mechanistically Interpretable Wearable-Based Sleep Staging},
  booktitle = {IEEE Annual Congress on Artificial Intelligence of Things (AIoT)},
  year      = {2025},
  note      = {In press}
}

@inproceedings{kadem2025human,
  title={Human-Clinical AI Agent Collaboration},
  author={Kadem, Mason and Al-Khazraji, Baraa},
  booktitle={Proceedings of the AAAI Symposium Series},
  volume={6},
  number={1},
  pages={255--264},
  year={2025}
}

@article{rudin2019stop,
  title={Stop Explaining Black Box Machine Learning Models for High Stakes Decisions and Use Interpretable Models Instead},
  author={Rudin, Cynthia},
  journal={Nature Machine Intelligence},
  volume={1},
  number={5},
  pages={206--215},
  year={2019},
  publisher={Nature Publishing Group}
}

@article{lundberg2020local,
  title={From local explanations to global understanding with explainable AI for trees},
  author={Lundberg, Scott M and Erion, Gabriel and Chen, Hugh and DeGrave, Alex and Prutkin, Jordan M and Katz, Ronit and Himmelfarb, Jonathan and Bansal, Navdeep and Lee, Su-In},
  journal={Nature Machine Intelligence},
  volume={2},
  number={1},
  pages={56--67},
  year={2020}
}

\end{document}